\documentclass[journal]{IEEEtran}
\usepackage[numbers,sort&compress]{natbib}

\ifCLASSINFOpdf
  \usepackage[pdftex]{graphicx}
\else
  \usepackage[dvips]{graphicx}
\fi

\usepackage{comment}
\usepackage{subfigure}

\usepackage[cmex10]{amsmath}

\usepackage{array}

\usepackage{hyperref}

\begin{document}
\title{Neuroevolution in Games:\\State of the Art and Open Challenges}

%\author{
%\IEEEauthorblockN{Sebastian Risi}
%\IEEEauthorblockA{Center for Computer Games Research\\
%IT University of Copenhagen\\
%Copenhagen, Denmark\\
%sebr@itu.dk}
%\and
%\IEEEauthorblockN{Julian Togelius}
%\IEEEauthorblockA{Department of Computer Science and Engineering\\ New York University\\
%New York, USA\\
%julian@togelius.com}
%}
\author{Sebastian Risi and Julian Togelius\thanks{SR is with the IT University of Copenhagen, Copenhagen, Denmark. JT is with the Department of Computer Science and Engineering, New York University, New York, USA. Emails: sebr@itu.dk, julian@togelius.com}}

\maketitle

\begin{abstract}
This paper surveys research on applying neuroevolution (NE) to games. In neuroevolution, artificial neural networks are trained through evolutionary algorithms, taking inspiration from the way biological brains evolved.  We analyse the application of NE in games along five different axes, which are the role NE is chosen to play in a game, the different types of neural networks used, the way these networks are evolved, how the fitness is determined and what type of input the network receives. The article also highlights important open research challenges in the field.

\end{abstract}
%\printfigures
\section{Introduction} 

The field of artificial and computational intelligence in games is now an established research field, but still growing and rapidly developing\footnote{There are now two dedicated conferences (IEEE Conference on Computational Intelligence and Games (CIG) and AAAI Conference on Artificial Intelligence and Interactive Digital Entertainment (AIIDE)) as well as a dedicated journal (IEEE Transactions on Computational Intelligence and AI in Games). Furthermore, work in this field is published in a number of conferences and journals in neighbouring fields.}. In this field, researchers study how to automate the playing, design, understanding or adaptation of games using a wide variety of methods drawn from computational intelligence (CI) and artificial intelligence (AI) \cite{miikkulainen:bridge06,miikkulainen:cigames06, chellapilla1999survey}. One of the more common techniques, which is applicable to a wide range of problems within this research field, is neuroevolution (NE) \cite{yao1999evolving, floreano2008neuroevolution}. Neuroevolution refers to the generation of artificial neural networks (their connection weights and/or topology) using evolutionary algorithms. This technique has been used successfully for tasks as diverse as robot control~\cite{nolfi2000evolutionary}, music generation~\cite{hoover2012generating}, modelling biological phenomena~\cite{clune2013evolutionary} and chip resource allocation~\cite{gomez2001neuro} among many others.
\begin{figure}[t]
\centering
\includegraphics[width=3.5in]{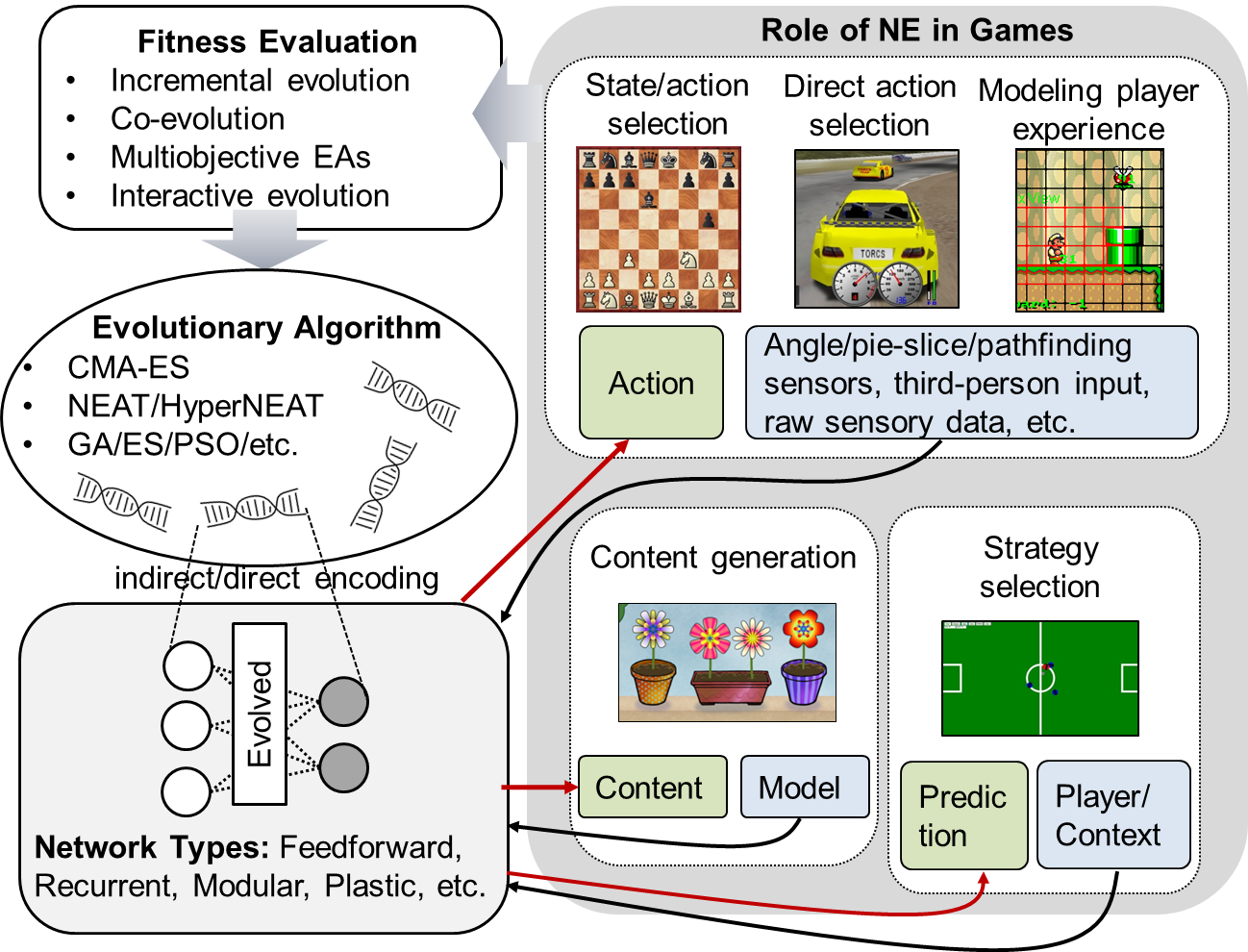}
\caption[]{\textbf{Neuroevolution in Games Overview}. An important distinction between NE approaches is the role that NE plays in a game, which is tightly coupled to  the input the evolved neural network receives (e.g.\ angle sensors) and what type of output it produces (e.g.\ a request to turn). NE's role  also directly influences the type of fitness evaluation. Different evolutionary algorithms support different network types and some methods can be more or less appropriate for different types of input representations. }
\label{fig:overview}
\end{figure} 

This paper surveys the use of neuroevolution in games (Figure~\ref{fig:overview}). A main motivation for writing it is that neuroevolution is an important method that has seen continued popularity since its inception around two decades ago, and that there are numerous existing applications in games and even more potential applications. The researcher or practitioner seeking to apply neuroevolution in a game application could therefore use a guide to the state of the art. Another main motivation is that games are excellent testbeds for neuroevolution research (and other AI research) that have many advantages over existing testbeds, such as mobile robotics. This paper is therefore meant to also be useful for the neuroevolution researcher seeking to use games as a testbed.

\subsection{Scope of this paper} 

In writing this paper, we have sought a broad and representative coverage of all kinds of neuroevolution to most kinds of games. While it is not possible to be exhaustive, we attempt to cover all of the main directions in this field and the most important papers. We only cover work where neuroevolution has in some way been applied to a game problem. By neuroevolution, we mean techniques where evolutionary computation or similar bio-inspired stochastic search/optimisation algorithms are applied to artificial neural networks (ANNs). We take an inclusive view of neural networks, including both weighted sums, self-organising maps  and multi-layer perceptrons, but not e.g.\ expression trees evolved with genetic programming. By games, we refer to games that people commonly play as games. This includes non-digital games (e.g.\ board games and card games) and digital games (e.g.\ arcade games, racing games, strategy games) but not purely abstract games such as prisoner's dilemma, robotics tasks or non-game benchmarks for reinforcement learning, such as pole balancing. We acknowledge that there are bound to be some gray areas, as no delineation can be absolutely sharp.

There are several other surveys available that cover larger topics or topics that intersect with the topic of this paper. This includes several surveys on CI/AI in games in general~\cite{yannakakis2014panorama,miikkulainen2006computational}, surveys on the use of evolutionary computation~\cite{lucas2006evolutionary} or machine learning~\cite{galway2008machine,munoz2013learning}, or surveys of approaches to particular research areas within the field such as pathfinding~\cite{botea2013pathfinding}, player modelling~\cite{yannakakis2013player} or procedural content generation~\cite{shaker2014procedural}.

\subsection{Structure of this paper} 

The next section gives an overview of neuroevolution, the main idea behind it, and the main motivation for employing a NE approach in games. Section~\ref{sec:role} details the first axes along which we analyse the role of neuroevolution in games, namely the type of role that the NE method is chosen to play. The different neural network types that are found in the NE games literature are reviewed in Section~\ref{sec:ann_types} and Section~\ref{sec:evolving_anns} then explains how they can be evolved. The different ways fitness can be evaluated in games are detailed in Section~\ref{sec:fitness}, followed by a review of different ANN input representations in Section~\ref{sec:input_rep}. Finally, Section~\ref{sec:outlook} highlights important open challenges in the field.

\section{Neuroevolution} 

The reasons for the wide applicability and lasting popularity of neuroevolution include that very many AI and control problems can be cast as optimization problems, where what is optimized is a general function approximator (such as a neural network). Another important reason for the attractiveness of this paradigm is that the method is grounded in biological metaphor and evolutionary theory. This section introduces the basic concepts and ideas behind neuroevolution. NE is motivated by the evolution of biological nervous systems and applies abstractions of natural evolution (i.e.\ \emph{evolutionary algorithms}) to generate artificial neural networks. ANNs are generally represented as networks composed of interconnected nodes (i.e.\ \emph{neurons}) that are able to compute values based on external inputs provided to the network. The ``behavior'' of an ANN is typically determined based on the architecture  of the network and the strength (i.e.\ \emph{weights}) of the connections between the neurons. 

Training an ANN to solve a computational problem involves finding suitable network parameters such as its topology and/or weights of the synaptic connections. The basic idea behind neuroevolution is to train the network with an evolutionary algorithm, which is a class of stochastic, population-based search methods inspired by Darwinian evolution. An important design choice in NE is the genetic representation (i.e.\ genotype) of the neural network that the combination and mutation operators manipulate. For example, one of the earliest and most straightforward ways to encode an ANN with a fixed topology (i.e.\ the topology of the network is determined by the user) is based on the concatenation  of the  numerical network weight values into a vector of real numbers.

\subsection{Basic Algorithm}
The basic NE algorithm works as follows. A \emph{population} of genotypes that encode ANNs is evolved to find a network (weight and/or topology) that can solve a computational problem. Typically each genotype is encoded into a neural network, which is then tested on a specific task for a certain amount of time. The performance or \emph{fitness} of this network is then recorded and once the fitness values for the genotypes in the current population are determined, a new population is generated by slightly  changing the ANN-encoding genotypes (mutation) or by combining multiple genotypes (cross-over). In general, genotypes with a higher fitness have a higher chance of being selected for reproduction and their offspring replaces genotypes with lower fitness values, thereby forming a new generation. This generational loop is typically repeated hundreds or thousands of times, in the hopes to finding  better and better performing networks. For a more complete review of NE see \citet{floreano2008neuroevolution}.

\subsection{Why Neuroevolution?}

\begin{figure*}
	\centering
	\subfigure[]{
		\includegraphics[height=0.158\linewidth]{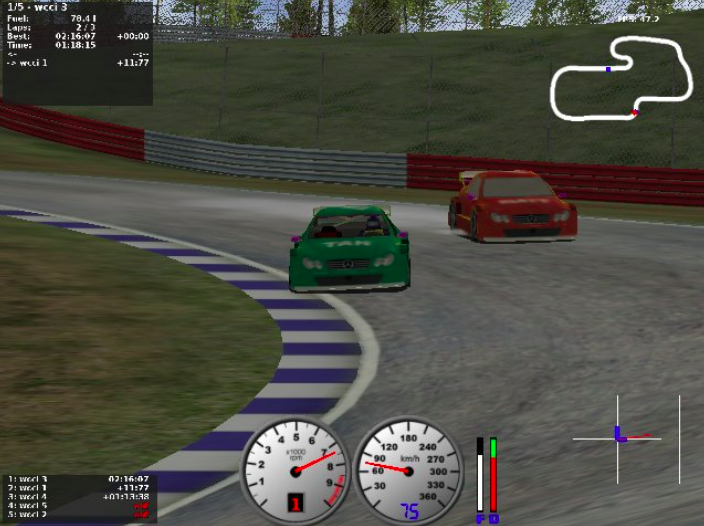}
}
	\subfigure[]{
		\includegraphics[height=0.158\linewidth]{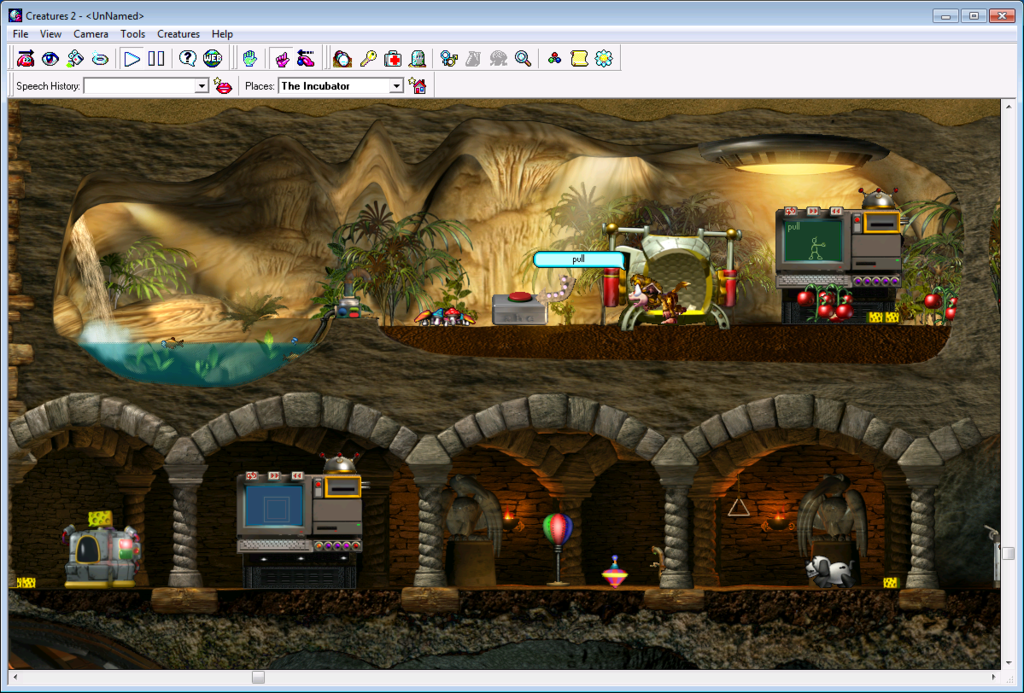}
		}
	\subfigure[]{
		\includegraphics[height=0.158\linewidth]{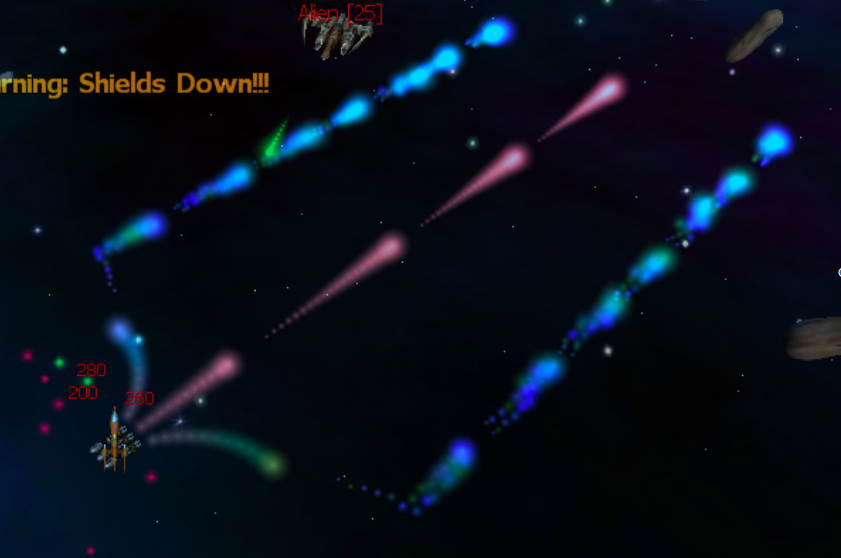}
		}
	\subfigure[]{
		\includegraphics[height=0.158\linewidth]{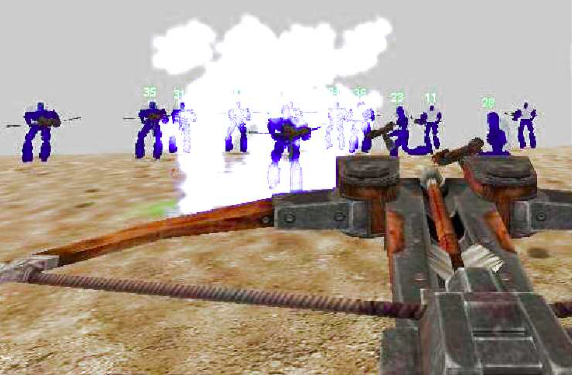}
	}
	\caption{{\bf Neuroevolution in Existing Games.} (a) NE is able to discover high-performing controllers for  racing games such as TORCS  \cite{cardamone2009line}. (b) NE has also been successfully applied to commercial games, such as Creatures~\cite{grand:agents97}. Additionally, NE enables new types of games such as GAR (c), in which players can interactively evolve particular weapons \cite{hastings2009automatic}, or NERO (d), in which players are able to evolve a team of robots and battle them against other players \cite{stanley2005real}.}
\label{fig:game_examples}
\end{figure*}

For each of the tasks that are described in this paper, there are other methods that could potentially be used. Game strategies could be learned by algorithms from the temporal difference learning family, player models could be learned with support vector machines, game content could be represented as constraint models and created with answer set programming, and so on. However, there are a number of reasons why neuroevolution is a good general method to apply for many of these tasks and an interesting method to study in all cases. Figure~\ref{fig:game_examples} shows selected examples of NE in existing games that highlight some of its unique benefits.

\subsubsection{Record-beating performance}
\label{sec:recordbeating}
For some problems, neuroevolution simply provides the best performance in competition with other learning methods (``performance'' is of course defined differently for different problems). This goes for example for various versions of pole balancing, a classic reinforcement learning problem, where the CoSyNE neuroevolution method convincingly beat all other methods for most problem parameters~\cite{gomez2008accelerated}. In particular, it found solutions to the problem using fewer tries than any other algorithm. The best performance on the Keepaway Soccer problem, another popular reinforcement learning benchmark, is also exhibited by a neuroevolution method~\cite{verbancsics2010evolving}. Neuroevolution can also perform very well on supervised learning tasks, as demonstrated by the performance of Evolino on several sequence prediction problems~\cite{schmidhuber2007training}. Here, a combination of neuroevolution and simple linear fitting could predict complex time-varying functions with lower error than any other method. In game domains, the winners of several game-based AI competitions are partly based on neuroevolution -- see for example the winners of the recent 2K BotPrize~\cite{schrum:cig11competition,schrum2012human} and Simulated Car Racing Championship~\cite{butz2009optimized}.

\subsubsection{Broad applicability}
Another benefit of NE is that it can be used for supervised, unsupervised and reinforcement learning tasks. Neuroevolution only requires  some sort of numeric evaluation of the quality of its candidate networks. In this respect, neuroevolution is similar to other kinds of reinforcement learning (RL) algorithms, such as those from the temporal difference (TD) family~\cite{sutton1998introduction}. However, if a dataset labelled with target values is provided NE can be used as a supervised learning algorithm similarly to how backpropagation is used. 

\subsubsection{Scalability} %Robustness}
Compared to many other types of reinforcement learning, especially algorithms from the TD family, neuroevolution seems to handle large action/state spaces very well, especially when used for direct action selection~\cite{parker2008neuro, parker2012neurovisual, floreano2004coevolution, TCIAIG13-mhauskn}.

%\cite{stanley2005real}.

\subsubsection{Diversity}
Neuroevolution can draw on the rich family of diversity-preservation methods (such as niching) and multiobjective methods that have been developed within the evolutionary computation community. This enables neuroevolution methods to achieve several forms of diversity among its results, and so deliver sets of meaningfully different strategies, controllers, models and/or content~\cite{agapitos2008generating,van2009robust}.

\subsubsection{Open-ended learning}

While NE can be used for RL in the same way that TD-learning can, one might argue that it could go beyond this relatively narrow formulation of reinforcement learning.
%n argument can be made that unlike those algorithms, neuroevolution incorporates a greater element of exploration and is less constrained.
In particular in cases where the topology of the network is evolved as well, neuroevolution could in principle support open-ended evolution, where behavior of arbitrary complexity and sophistication could emerge. Concretely, neuroevolution algorithms often search in a larger space than TD-based algorithms do.

\subsubsection{Enables new kinds of games}

New video games such as \emph{Galactic Arms Race} (GAR; \cite{hastings2009automatic}), in which the player can interactively evolve particular weapons, NERO \cite{stanley2005real}, which allows the player to evolve a team of robots and battle them against other players, or the \emph{Petalz} video game \cite{risi2012combining,risi12015petalz}, in which the player can breed an unlimited variety of virtual flowers, would be difficult to realize with traditional learning methods. Evolutionary computation here provides unique affordances for game design, and some designs rely specifically on neuroevolution. In the case of games like Petalz and GAR, the games rely on the continuous  complexification of the produced content, which is (naturally) supported and an integral part of certain NE methods. Additionally, in games like Petalz whose core game mechanic is the breeding of new flower types, employing evolutionary methods is a natural choice. An important example of a commercial game series that offers novel gameplay based on NE is \emph{Creatures} \cite{grand:agents97}. In the first Creatures game, which was created in the mid-1990s, the player can breed and raise virtual pets called Norns, and teach them to survive in their environment. In contrast to most other games, the adaptive ANNs controlling the pets actually allow them to learn new behaviors guided by the player. 

The fact that neuroevolution methods facilitate open-ended learning by incorporating a greater element of exploration, as highlighted in the previous section, also makes NE directly applicable to support new kinds of games.

\subsection{Why not Neuroevolution?}

While neuroevolution has multiple attractive properties, it is not a panacea and there are several reasons why other methods might be more suitable to particular problems. The most important one is that the evolved neural networks tend to have ``black box'' characteristics, meaning that a human cannot easily work out what they do by looking at them. This is a problem for game development and in particular quality assurance, as it becomes very hard to ``debug'' learned behavior. A problem with using neuroevolution online is that it is very hard to predict exactly what kind of behavior will be learned, something which can clash with the traditional design principles of commercial games.

\section{Role of neuroevolution}
\label{sec:role}

The role that neuroevolution is chosen to play is the first of several axes along which NE's usage is analysed in this paper. According to our survey of the literature, in the vast majority of cases neuroevolution is used to learn to play a game or control an NPC in a game. The neural network can here have one of two roles: to {\it evaluate the value of states or actions}, so that some other algorithm can choose which action to take, or to {\it directly select actions} to take in a given state. But there are also other uses for neuroevolution. \emph{Procedural content generation} (PCG) is an active area of research in game AI, and here evolvable neural networks can be used to {\it represent content}. Finally, neuroevolution can also {\it predict the experience or preferences} of players. Table~\ref{tab-examples} summarizes the usage of NE in a variety of different games.

%selected examples of how NE is successfully applied to a variety of different games. 

This section will survey the use of neuroevolution in games in each role; each section will be ordered according to game genre. 

%performance alone = PA
%(Number of features)
%Car racing floreano
%Go Roving Eye
%Modular MO
%Selected Examples of Neuroevolution Usage in Games.
%CNE = conventional neuroevolution (e.g.\ genetic algorithm + user-defined topology)

\begin{table*}
\begin{center}
\caption{\bf{The Role of Neuroevolution in Selected Games.} \normalfont ES = evolutionary strategy, GA = genetic algorithm, MLP = multi-layer perceptron, MO = multiobjective, TP = third-person (input not tied to a specific frame of reference, e.g.\ number edible ghosts) , UD =  user-defined network topology, PA = performance alone}
\label{tab-examples}
\begin{tabular}{|l|l|l|l|l|l|}
  \hline
  NE Role & Game & ANN Type & NE Methods & Fitness Evaluation & Input Representation\\
  (Section~\ref{sec:role}) &  & (Section~\ref{sec:ann_types})  & (Section~\ref{sec:evolving_anns})  & (Section~\ref{sec:fitness}) & (Section~\ref{sec:input_rep}) \\
  \hline\hline
  State/action & {\bf Checkers}~\cite{fogel2004self} & MLP & UD, GA & Coevolution & TP (piece type)\\
              evaluation & {\bf Chess}~\cite{fogel2004self}& MLP & UD, GA & PA (positional values) & TP (piece type) \\
                          & {\bf Othello}~\cite{moriarty1994evolving} & MLP & Marker-based \cite{fullmer1992evolving} & Cooperative coevolution & TP (piece type) \\
                          & {\bf Go (7$\times$7)}~\cite{gauci2010indirect} & CPPN (MLP) & HyperNEAT & PA (score+board size) &  TP (piece type)\\
  & {\bf Ms. Pac-Man}~\cite{lucas2005evolving} & MLP & UD, ES & PA (average score) & Path-finding\\
                          & {\bf Simulated Car Racing}~\cite{lucas2007point} & MLP & UD, ES & PA (waypoints visited) & Speed, pos, waypoints\\
  \hline
  Direct action & {\bf Quake II} \cite{parker2008neuro, parker2012neurovisual} & MLP & UD, GA & PA (kill count)&  Visual Input (14$\times$2)\\
  selection & {\bf Unreal Tournament} \cite{van2009hierarchical} &   Recurrent, LSTM & UD, GA, NSGA-II & MO (damage\&accuracy) &  Pie-slice, way point, etc.\\

 & {\bf Go (7$\times$7)}~\cite{stanley:gecco04} & MLP & NEAT & Transfer Learning & Roving Eye (3$\times$3)\\
  & {\bf Simulated Car Racing}~\cite{togelius2006evolving} & MLP & UD, ES & Incremental Evolution & Rangefinders, waypoints\\
  & {\bf Keepaway Soccer}~\cite{AAMAS07-taylor} & MLP & NEAT & Transfer Learning  & Distances \\
  & {\bf Battle Domain}~\cite{schrum2008constructing} & MLP & NEAT, NSGA-II & MO+Incremental & Angle,  straight line\\
  & {\bf NERO}~\cite{stanley2005real}& MLP & NEAT & Interactive Evolution & Rangefinders, pie-slice\\
  & {\bf Ms. Pac-Man}~\cite{schrum2:gecco14} & Modular MLP & NEAT, NSGA-II & MO (pills\&ghosts eaten) & Path-finding\\
  & {\bf Simulated Car Racing}~\cite{floreano2004coevolution} & MLP & UD, GA & PA (distance) & Roving Eye (5$\times$5) \\
  & {\bf Atari}~\cite{TCIAIG13-mhauskn} & CPPN (MLP) & HyperNEAT & PA (game score) & Raw input (16$\times$21)\\
    & {\bf Creatures}~\cite{grand:agents97} & Modular MLP & GA & Interactive Evolution & TP (e.g.\ type of object)\\
  \hline
  Selection between   & {\bf Keepaway Soccer}~\cite{whiteson2010critical,whiteson2007empirical} & MLP & NEAT & PA (hold time) & Angle and distance \\
   strategies & {\bf EvoCommander}~\cite{evocommander2015} & MLP & NEAT & Interactive Evolution & Pie-slice, rangefinder \\
  
  \hline
  Modelling opponent  & {\bf Texas Hold'em Poker}~\cite{lockett:2008} & MLP & NEAT & PA (\%hands won) & TP (e.g.\ size of pot,\\
  strategy & & & & & cost of a bet, etc.)\\
  \hline
  Content generation & {\bf GAR}~\cite{hastings2009automatic} & CPPN (MLP) & NEAT & Interactive Evolution & Model \\
    & {\bf Petalz}~\cite{risi12015petalz} & CPPN (MLP) & NEAT & Interactive Evolution & Model \\
  \hline
  Modelling player  & {\bf Super Mario Bros} \cite{pedersen2010modeling} & MLP, Perceptron & UD, GA & PA (player preference) & TP (e.g.\ gap width, \\
  experience & & & & & number deaths, etc.)\\
  \hline
\end{tabular}
\end{center}
\end{table*}

\subsection{State/action evaluation}
\label{sec:action_eval}

The historically earliest and possibly most widespread use of
neuroevolution in games is to estimate the value of board positions for agents playing classical board games. In these examples, a tree search algorithm of some kind (such as Minimax with alpha-beta pruning and/or other modifications) is used to search the space of future moves and counter-moves. The role of the evolved neural network here is to evaluate the quality of a hypothetical future board state (the nodes of the search tree) and assign a numerical value to it. This quality value is supposed to be related to how close you are to winning the game, so that the rational course of action for a game-playing agent is to choose actions which would lead to board states with higher values. In most cases, the neural network is evolved using a fitness function based on the win rate of the game against some opponent(s), but there are also examples where the fitness function is based on something else, e.g.\ human playing style.

The most well-studied game in the history of AI is probably Chess. However, relatively few researchers have attempted to apply neuroevolution to this problem.  Fogel et al. devised an architecture that was able to learn to play above master level solely by playing against itself~\cite{fogel2004self}.

There has also been plenty of work done on evolving board evaluators for the related, though somewhat simpler, board game Checkers. This was the focus of a popular science book by Fogel in 2001~\cite{fogel2001blondie24}, which chronicled several years' work on evolving Checkers players~\cite{chellapilla1999evolving,chellapilla2001evolving}. The evolved player, called Blondie24, managed to learn to play at a human master level. However, interest in Checkers as a benchmark game waned after the game was solved in 2007~\cite{schaeffer2007checkers}. It has been shown that even very simple networks can be evolved to play Checkers well~\cite{hughes2003piece}.

Another classic board game that has been used as a testbed for neuroevolution is Othello (also called Reversi). Moriarty and Miikkulainen used cooperative coevolution to evolve board evaluators for Othello \cite{moriarty1994evolving, moriarty1995discovering}. Chong et al. evolved board evaluators based on convolutional multi-layer neural networks that captured aspects of board geometry~\cite{chong2005observing}. In contrast, Lucas and Runarsson later used evolution strategies to learn position evaluators that were remarkably effective though they were simply weighted piece counters~\cite{lucas2006temporal}. They also compared evolution to temporal difference learning for the same problem, a topic we will return to in Section~\ref{section:comparing}. Later work by the same authors showed that the n-tuple network, an evolvable neural network based on sampling combinations of board positions, could learn very good state evaluators~\cite{lucas2014preference}; n-tuple networks were also shown to perform better than other neural architectures on playing Checkers~\cite{al2011investigating}.

Go has in recent years become the focus of much research in AI game-playing. This is because this ancient Asian board game, despite having very simple rules, is very hard for computers to play proficiently. Until a few years ago, the best players barely reached intermediate human play level. The combination of Minimax-style tree search and board state evaluation that has worked so well for games such as Chess, Checkers and Othello largely breaks down when applied to Go, partly because Go has much higher branching factor than those games, and partly because it is very hard to calculate good estimates of the value of a Go board. Early attempts to evolve Go board state evaluators used fairly standard neural network architectures~\cite{lubberts:geccoworkshop01,richards:icga97}. It seems that such methods can do well on small-board Go (e.g.\ $5\times 5$) but fail to scale up to larger board sizes~\cite{runarsson2005coevolution}. Gauci et al. have attempted to address this by using HyperNEAT (see Section~\ref{sec:evolving_anns}), a neural network architecture specially designed to exploit geometric regularities to learn to play Go~\cite{gauci2010indirect}; Schaul and Schmidhuber have tried to address the same issue using recurrent convolutional networks~\cite{schaul2009scalable}. It should be noted that neuroevolution is not currently competitive with the state of the art in this domain; the state of the art for Go is Monte Carlo Tree Search (MCTS), a statistical tree search technique which has finally allowed Go-playing AI to reach advanced human level~\cite{browne2012survey}.

But the use of neural networks as state value evaluators goes beyond board games. In fact any game can be played using a state value evaluation function, given that it is possible to predict which future states actions lead to, and that which there is a reasonably small number of actions. The principle is the same as for board games: search the tree of possible future actions and evaluate the resulting states, choosing the action that leads to the highest-valued state. This general method works even in the case of non-adversarial games, such as typical single-player arcade games, in which case we are building a Max-tree rather than a Minimax-tree. Further, sometimes good results can be achieved with very shallow searches, such as a one-ply search where only the resulting state after taking a single action is evaluated. A good example of this is Lucas' work on evolving state evaluators for Ms. Pac-Man~\cite{lucas2005evolving} which has inspired a number of further studies~\cite{schrum2:gecco14,cardona2013competitive}. As Pac-Man never has more than four possible actions, this makes it possible to search more than one ply, as can be seen for example in the work of Borg Cardona et al.~\cite{cardona2013competitive}. 

It is possible to evolve evaluators not only for states, but also for future actions. The neural network would in this case take the current state and potential future action as input, and return a value for that action. Control happens through enumerating over all actions, and choosing the one with the highest value. Lucas and Togelius compared a number of ways that neuroevolution could be used to control cars in a simple car racing game (with discrete action space), including evolving state evaluators and evolving action evaluators~\cite{lucas2007point}. It was found that evolving state evaluators led to significantly higher performance than evolving action evaluators.

\subsection{Direct action selection}
\label{sec:direct_action}

It is not always possible to play a game (or control a game NPC) through evaluating potential future states or actions. For example, there might be too many actions from any given state to be effectively  enumerable -- a game like {\it Civilization} has an astronomical number of possible actions, and even a one-ply search would be computationally prohibitively expensive. (Though see Branavan et al.'s work on learning of action evaluators in {\it Civilization II} through a hybrid non-evolutionary method~\cite{branavan2011non}.) Further, there are cases where you do not even have a reliable method of predicting which state a future action would lead to, such as when learning a player for a game which you do not have the source code for, when the forward model for calculating this state would be too computationally expensive, or when the forward model is stochastic.

In such cases, neural networks can still play games and control NPCs through direct action selection. This means that the neural network receives a description of the current state (or some observation of the state) as an input and outputs which action to take. The output could take different forms, for example the neural network might have one output for each action, or continuous outputs that define some action space. In many cases, the output dimensions of the network mirror or are similar to the buttons and sticks on a game controller that would be used by a human to play the game. (Note that the state evaluators and action evaluators discussed in the previous section also perform action selection, but in an \emph{indirect} way.)

In a number of experiments with a simple car racing game, Togelius and Lucas evolved networks that drove the cars using a number of sensors as inputs, and outputs for steering and acceleration/braking~\cite{togelius2005evolving,togelius2006evolving,togelius2007multi}. It was shown that these networks could drive as least as well as a human player, and be incrementally evolved to proficiently drive on a wide variety of tracks. Using competitive coevolution, a variety of interestingly different playing styles could be generated. However, it was also shown that for a version of this problem, evolving networks to evaluate states gave superior driving performance compared to evolving networks to directly select actions~\cite{lucas2007point}. This work on evolving car racing controllers spawned a series of competitions on simulated car racing; the first year's competition built on the same simple racing game~\cite{togelius20082007}, which was exchanged for the more sophisticated racing game {\it TORCS} for the remaining years~\cite{loiacono20102009,loiacono2008wcci}. Several of the high-performing submissions were based on neuroevolution, usually in the role of action selector~\cite{cardamone2009line}, but sometimes in more auxiliary roles~\cite{butz2009optimized}.

Another genre where evolved neural networks have been used in the role of action selectors is first-person shooter (FPS) games. Here, the work of Parker and Bryant on evolving bots for the popular 90's FPS {\it Quake II} is a good example~\cite{parker2008neuro,parker2012neurovisual}. In this case, the neural networks have different outputs for turning left/right, moving forward/backward and shooting. The FPS game which has been used most frequently for AI/CI experimentation is probably Unreal Tournament, as it has an accessible interface (called Pogamut~\cite{gemrot2009pogamut}) and has been used in a popular competition, the {\it 2K BotPrize}~\cite{hingston2010new}. van Hoorn et al. evolved layered controllers consisting of several neural networks that implemented different components of good game-playing behavior: path following exploration and shooting~\cite{van2009hierarchical}. The different networks output direct control signals corresponding to movement directions and shooting, and were able to override each other according to a fixed hierarchy. Schrum et al. evolved neural networks to behave in a human-like manner using the same testbed~\cite{schrum:cig11competition}. Another game with strong similarities to first-person shooter in terms of the capabilities of the agents (though a completely different focus for its gameplay) is the experimental game NERO, which sees each of its agents controlled by its own neural network evolved through a version of the popular NE method NEAT~\cite{stanley2005real}. NEAT is explained in more detail in Section~\ref{sec:evolving_anns}.

In two-dimensional platform games, the number of actions possible to take at any given moment is typically very limited. This partly goes back to the iconic platform games, such as Nintendo's {\it Super Mario Bros}, being invented in a time when common game platforms had limited input possibilities. In Super Mario Bros, the action space is any combination of the buttons on the original Nintendo controller: up, down, left, right, A and B. Togelius et al. evolved neural networks to play the Infinite Mario Bros clone of Super Mario Bros using networks which simply had one output for each of the buttons \cite{togelius2009super}. The objective, which met with only limited success, was simply to get as far as possible on a number of levels. A study by Ortega et al. subsequently used neuroevolution in the same role but with a different objective, evolving networks to mimic the behaviour of human players~\cite{ortega2013imitating}.

Whereas it might seem that when using neural networks for playing board games the networks should be used as state or action evaluators, there have been experiments with using neuroevolution in the direct action selection role for board games. This might mean having one network with an output for each board position, so that the action taken is the possible action which is associated with the highest-valued output; this is the case in the Checkers-playing network of Gauci et al.~\cite{gauci2010autonomous}. A more exotic variant is Stanley and Miikkulainen's ``roving eye'', which plays Go by self-directedly traversing the board and stopping where it thinks the next stone should be placed~\cite{stanley:gecco04}. These attempts are generally not competitive performance-wise with architectures where the neural network evaluates states, and are done in order to test new network types or controller concepts.

Finally, there has been work on developing architectures for neural networks that can be evolved to play any of a large number of games. One example is Togelius and Schmidhuber's neural networks for learning to play any of a series of discrete 2D games from a space defined in a simple game description language~\cite{togelius2008experiment} (this work later inspired the General Video Game Playing Competition; see Section~\ref{gvgp}). More recently, Hausknecht et al. evolved neural networks of different types to play original Atari 2600 games using the Atari Learning Environment~\cite{TCIAIG13-mhauskn,hausknecht:gecco12}. The network outputs here were simply mapped to the Atari controller; performance varied greatly between different games, with evolved networks learning to play some games at master levels and many others barely at all.

\subsection{Selection between strategies}

For some games, it makes no sense to control the primitive actions of a player character, but rather to choose between one of several strategies to be played for a short time span (longer than a single time step). An example is the \emph{Keepaway Soccer} task, which is a reinforcement learning benchmark with strongly game-like qualities as it is extracted from Robocup, a robot football tournament. Every time an agent has the ball, it selects one out of three static macro-actions, which is played out until the next ball possession \cite{whiteson2010critical,whiteson2007empirical}. Whiteson et al. evolved neural networks for this task using NEAT with good results, but found that hybridising with other forms of reinforcement learning worked even better~\cite{whiteson2007empirical}.

\subsection{Modelling opponent strategy}

In many cases, a player needs to be able to predict how its opponent will act in order to play well. Neuroevolution can be used in the specific role of predicting opponent strategy, as part of a player whose other parts might or might not be based on neuroevolution. Lockett and Miikkulainen evolved networks that could predict the other player's strategy in Texas Hold'em Poker, increasing the win rate of agents that used the model~\cite{lockett:2008}.

\subsection{Content generation}
\label{sec:content_gen}

Recently interest has increased in a field called procedural content generation (PCG) \cite{shaker2014procedural}, in which parts of a game are created algorithmically rather than being hand-designed. Especially stochastic search methods, such as evolutionary algorithms, have shown promise in generating various types of content, from game levels to weapons and even the rules of the game itself; evolutionary approaches to PCG go by the name of {\it search-based} PCG~\cite{togelius2011searchbased}. 

In several applications, the content has been represented as a neural network. In these applications, the output characteristic (or decision surface) of the network is in some way interpreted as the shape, hull or other characteristic of the content artifact. The capacity for neural networks to encode smooth surfaces of essentially infinite resolution makes them very suitable for compactly encoding structures with continuous shapes. In particular, a type of neural network called a Compositional Pattern Producing Network (CPPN~\cite{stanley2007compositional}) is of interest. CPPNs are a variation of ANNs that differ in the type of activation functions they contain and also the way they are applied. CPPNs are described in more detail in Section~\ref{sec:evolving_anns}. 
Examples of CPPNs used in content generation include the Galactic Arms Race video game (GAR~\cite{hastings2009automatic}), in which the player can evolve CPPN-generated weapons, and the Petalz video game \cite{risi2012combining,risi2014automatically,risi12015petalz}, in which the player can interactively evolve CPPN-encoded flowers. In another demonstration, \citet{liapis2012adapting} evolved CPPNs to generate visually pleasing two-dimensional spaceships, which can be used in space shooter games. In more recent work, the authors augmented their system to allow the autonomous creation of spaceships based on an evolving interestingness criterion \cite{liapis2013delenox}.

\subsection{Modelling player experience}

Many types of neural networks (including the standard multi-layer perceptron) are general function approximators, i.e.\ they can approximate any function to a given accuracy given a sufficiently large number of neurons. This is a reason that neural networks are very popular in supervised learning applications of various types. Most often, neural networks used for supervised learning are trained with some version of  backpropagation; however, in some cases neuroevolution provides superior performance even for supervised learning tasks. This is in particular the case with preference learning, where the task is to learn to predict the ordering between instances in a data set.

One prominent use of preference learning, in particular preference learning through neuroevolution, is player experience modeling. Pedersen et al. trained neural networks to predict which level a player would prefer in a clone of Super Mario Bros~\cite{pedersen2010modeling}. The dataset was collected through letting hundreds of players play through pairs of different levels, recording the playing session of each player as well as the player's choice of which level in the pair was most challenging, frustrating and engaging. A standard multilayer perceptron was then given information on the player's playing style and features of both levels as inputs; the output was which of the two levels was preferred. Training these networks with neuroevolution, it was found that player preference could be predicted with up to 91\% accuracy. Shaker et al. later used these models to evolve personalised levels for this Super Mario Bros game, through simply searching the parameter space for levels that maximise particular aspects of player experience~\cite{shaker2010towards}.

\section{Neural Network Types} 
\label{sec:ann_types}

A variety of different neural network types can be found in the literature, and it is important to note that the type of ANN can significantly influence its learning abilities. The simplest neural networks are \emph{feedforward}, which means that the information travels directly from the input nodes through the hidden nodes to the output nodes. In feedforward networks there are no synaptic connections that form loops or cycles. The simplest feedforward network is the perceptron, which is an ANN without any hidden nodes (only direct feed-forward connections between the inputs and the outputs), and where the activation function is either the simple step function or the identity function (in the latter case the perceptron is simply a weighted sum of its inputs)~\cite{rosenblatt65perceptron}. The classic perceptron without hidden nodes can only learn to correctly solve linearly separable problems, limiting its use for both pattern classification and control. The more complex multilayer perceptron (MLP) architectures with hidden nodes and typically sigmoid activation functions can also learn problems that are not linearly separable, and in fact approximate any function to a given accuracy given a suitable large number of hidden neurons~\cite{hornik1989multilayer}. While this capability is mostly theoretical, the greater representational capacity of MLPs over standard perceptrons can readily be seen when applying neuroevolution to NPC control in games. For example, \citet{lucas2005evolving}  compared the performance of single and multi-layer perceptrons in a simplified version of the Ms. Pac-Man game and showed that the MLP reached a significantly higher performance.

In \emph{recurrent networks}, the network can form loops and cycles, which allows information to also be propagated from later layers back to earlier layers. Because of these directed cycles the network can create and maintain an internal state that allows it to exhibit dynamic temporal properties and keep a memory of past events when applied to control problems. Simply put, recurrent networks can have memory, whereas static feedforward networks live in an eternal present. Being able to base your actions on the past as well as the present has obvious benefits for game-playing. In particular, the issue of sensory aliasing arises when the agent has imperfect information (true for any game in which not all of the game world is on screen at the same time), so that different states which are very different look the same to the agent~\cite{ziemke1999remembering}. Domains with this property are also called \emph{non-Markovian} tasks, meaning that the state of the environment is not fully observable by the agent \citep{sondik1978optimal}. Recurrent and non-recurrent networks have been shown to perform very similarly in domains such as platform game playing \cite{togelius2009super}, while recurrent networks have been shown to consistently outperform feedforward on a car racing task~\cite{van2009robust}. 

While most NE approaches in games focus on evolving monolithic neural networks, {\it modular} networks have recently shown superior performance in some domains \cite{schrum2:gecco14}. A modular network is composed of a number of individual neural networks (modules), in which the modules are normally responsible for a specific sub-function of the overall system.  For example, Schrum has recently shown that in the Pac-Man domain a modular network extension to NEAT performs better than the standard version which evolves monolithic (non-modular) networks. The likely explanation is that it is easier to evolve multimodal behavior with modular architectures~\cite{schrum2:gecco14}. That experiment featured modules which were initially undifferentiated but whose role was chosen by evolution. Explicitly functionally differentiated modules can also work well when the task lends itself to easy functional decomposition. van Hoorn et al. evolved separate neural modules for path-following, shooting and exploration; these modules were arranged in a hierarchical architecture, an arrangement which was shown to perform much better than a monolithic network~\cite{van2009hierarchical}. Outside of games, there has been research on ways of encouraging neural networks to evolve modular structures~\cite{clune2013evolutionary}, and also on evolving ensembles of neural networks~\cite{yao2008evolving} where the different networks have complementary fitness functions~\cite{abbass2003pareto}.

While most types of neural networks do not change their connection weights after initial training, {\it plastic} neural networks can change their connection weights at any time in response to the activation pattern of the network. This enables a form of longer-term learning than what is practically possible using recurrent networks. Plastic networks have been evolved to solve robot tasks in unpredictable environments which the controller does not initially know characteristics of~\cite{floreano1996evolution}, however they have not yet been applied to games to our best knowledge. It seems that plastic networks would be a promising avenue for learning controllers that can play multiple games or multiple versions of the same game. Another adaptive architecture that has shown promising results in learning from experience, is a \emph{Long-Short Term Memory} (LSTM)
network \cite{gers2001lstm}. Instead of changing connection weights, LSTM networks can learn by remembering values for an arbitrary amount of time. We will return to the topic of life-time learning when we discuss important open challenges in Section~\ref{sec:outlook}.

\section{Evolving Neural Networks}
\label{sec:evolving_anns}

A large number of different evolutionary algorithms have been applied to NE, including genetic algorithms, evolution strategies and evolutionary programming~\cite{yao1999evolving}; in addition, other stochastic search/optimisation methods like particle swarm optimisation have also been used~\cite{eberhart2001particle}.
The method for evolving a neural network is tightly coupled to its genotypic representation. The earliest NE methods only evolved the weights of a network with a fixed user-defined topology and in the simplest \emph{direct representation}, each connection is encoded by a single real value. This encoding, sometimes called \emph{conventional neuroevolution} (CNE),  allows the whole network to be described as a concatenation of these values, i.e.\ a vector of real numbers. For example, both \citet{cardona2013competitive} and \citet{lucas2005evolving} directly evolved the weights of a MLP for Ms. Pac Man. Being able to represent a whole network as a vector of real numbers allows the use of any evolutionary algorithm which works on vectors of real numbers, including highly efficient algorithms like the Covariance Matrix Adaptation Evolution Strategy (CMA-ES)~\cite{hansen2001completely,igel2003neuroevolution}.

A significant drawback of the fixed topology approach is that the user has to choose the appropriate  topology and number of hidden nodes a priori. Because the topology of a network can significantly affect its performance~\cite{igel2003neuroevolution,togelius2007multi,togelius2008countering,togelius2009super}, more sophisticated approaches evolve the network topology together with the weights. An example of such a method is NeuroEvolution of Augmenting Topologies (NEAT; \cite{stanley:ec02}), which has shown that also evolving the topology together with the connection weights often outperforms approaches that evolve the weights alone. NEAT and similar methods also allow the evolution of recurrent networks that can solve difficult non-Markovian control problems (examples outside of games include pole balancing~\cite{gomez2008accelerated} and tying knots~\cite{mayer2008system}). NEAT and its variants\footnote{NEAT now also runs under the Unity Game Engine: \url{https://github.com/lordjesus/UnityNEAT}. A complete list of NEAT implementations can be found here: \url{http://bit.ly/1J1II8O}} have been applied successfully  to a variety of different game domains, from controlling a simulated car in The Open Car Racing Simulator (TORCS)~\cite{cardamone2009evolving} or a team of robots in the NERO game \cite{stanley2005real} to playing Ms.~Pac-Man~\cite{schrum2:gecco14} or Unreal Tournament~\cite{schrum:cig11competition}. Of particular interest in the context of this paper is an extension to NEAT that has been developed to increase its performance in strategic-decision making problems \cite{kohl2009evolving}. \citet{kohl2009evolving} demonstrate that by adding a new topological mutation that adds neurons with local fields to the network (RBF-NEAT) and a node mutation that biases the network architecture to cascade type of structures (Cascade-NEAT), the approach is able to more easily solve \emph{fractured} strategic-decision task such as keepaway soccer. Fracture is defined by the authors as a ``highly discontinuous mapping between states and optimal actions'' within an ANN. It is likely that the more complex the game, the more it will have a fractured decision space.

%a variation of NEAT called \emph{Cascade-NEAT}, performs better 
%interesting NEAT extension was proposed by .    that strategic-decision making   
% 4-versus-2 keepaway soccer problem
%TODO   (These results might also be explained by the underlying fractured decision space of %such a mapping~\cite{kohl2009evolving}; see Section~\ref{sec:evolving_anns}). 

Both the fixed-topology and topology-evolving approaches above use a direct encoding, meaning that each connection is encoded separately as a real number.
Direct encoding approaches employ a one-to-one mapping between the parameter values of the network (e.g.\ the weights and neurons) and its genetic representation. One disadvantage of this approach is that parts of the solution that are similar must be discovered separately  by evolution. Therefore, interest has increased in recent years in \emph{indirect encodings}, which allow the reuse of information to encode the final network and thus very compact genetic representations (i.e.\ a high number of synaptic connections can be encoded by considerably fewer parameters in the corresponding genotype).

A promising indirect encoding that has been employed successfully in a variety of NE-based games are CPPNs \cite{stanley2007compositional} (see Section~\ref{sec:content_gen}). CPPNs are a variation of ANNs that differ in the type of activation functions they contain and also the way they are applied. While ANNs are traditionally employed as controllers, CPPNs often function as pattern-generators, and have shown promise in the context of procedurally generated content. Additionally, while ANNs typically only contain sigmoid activation functions, CPPNs can include these but also a variety of other activation functions like sine (to create repeating patterns) or Gaussian (to create symmetric patterns). Importantly, because CPPNs are variations of ANNs they can also be evolved by the NEAT algorithm~\cite{stanley2009hypercube}. 

While CPPNs were initially designed to produce two-dimensional patterns (e.g.\ images), they have also been extended to evolve indirectly encoded ANNs. The main idea in this approach, called \emph{HyperNEAT}~\cite{stanley2009hypercube} is to exploit geometric domain properties to compactly describe the connectivity pattern of a large-scale ANN. For example, in a board game like chess or checkers adjacency relationships or symmetries play an important role and understanding these regularities can enable an algorithm to learn general tactics instead of specific actions associated with a single board piece. 

\citet{TCIAIG13-mhauskn} recently showed that HyperNEAT's ability to encode large scale ANNs enables it to directly learn Atari 2600 games from raw game screen data, outperforming human high scores in three games. This advance is exciting from a game perspective since 
it takes a step towards \emph{general} game playing that is independent from a game specific input representation. We will return to this topic in Section~\ref{sec:input_rep}.

Another form of indirect encodings are \emph{developmental approaches}. In a promising demonstration of this approach, \citet{khan2014search} evolved a single developing and more complex neuron that was able to play checkers and beat a Minimax-based program. In contrast to the aforementioned indirect encodings like CPPNs, the neuron in Khan and Miller's work grows (i.e.\ develops) new synaptic connections while the game is being played. This growth process is based on an evolved genetic program that processes inputs from the board.

\section{Fitness Evaluation}
\label{sec:fitness}

The fitness of a board value estimator or ANN controlled NPC is traditionally determined based on its performance in a particular domain. This might translate to the score it achieves in a single player game or how many levels it manages to complete, or how many opponents it can beat or which ranking it can achieve in a competitive multiplayer game. Because many games are non-deterministic leading to \emph{noisy fitness evaluation} (meaning that the same controller can score differently when evaluated several times on the same problem), evaluating the performance of a controller in multiple independent plays and averaging the resulting scores can be beneficial~\cite{runarsson2005coevolution,togelius2005evolving}. In many cases it is possible to evolve good behaviour using a straightforward fitness function like the score of the game, which has the advantage that such fitness functions can typically not be ``exploited'' in the way composite fitness functions sometimes can (the only way to achieve a high score is to play the game well).

However, if the problem becomes too complex, it can be difficult to evolve the necessary behaviors directly. In such cases it can be helpful to learn them incrementally, by first starting with a simpler task and gradually making them more complex; this is usually called \emph{staging} and/or \emph{incremental evolution}~\cite{gomez1997incremental}. For example, \citet{togelius2006evolving} evolved controllers for simulated car racing with such an incremental evolution approach. In their setup, each car is first evaluated on one track and each time the population reaches a sufficiently high fitness more challenging tracks are added to the evaluation and fitness averaged. The authors showed that by following such an incremental evolutionary approach, neurocontrollers with general driving skills evolve. No such general driving skills were observed when a controller was evaluated on all tracks simultaneously. Incremental evolution can also be combined with modularisation  of neural networks so that each time a new fitness function is added, a new neural network module is added to the mix; this could allow evolution to solve problems where the acquisition of a new competence conflicts with an existing competence~\cite{van2009hierarchical}.

A form of training that can be viewed as a more radical version of incremental evolution is \emph{transfer learning}. The goal of transfer learning is to accelerate the learning of a target task through knowledge gained during learning of a different but related source task. For example, \citet{AAMAS07-taylor} showed that transfer learning can significantly speed up learning through NEAT from 3~vs.~2 to 4~vs.~3 robot soccer Keepaway. In a demonstration of applying transfer learning to two different but related games, \citet{cardamone2011transfer} transferred a car racing controller evolved in TORCS to another open-source racing game called VDrift. By exploiting knowledge gained in TORCS, evolution was able to adapt the pre-existing model to the new game quicker than when the authors tried to evolve a controller in VDrift from scratch.

While incremental evolutionary approaches allow NE to solve more complex problems, designing  a ``good'' fitness function can be a challenging problem, especially if the tasks become more and more complex. In particular, this is the case for competitive games where a good and reliable opponent AI is not available, and for games where an evolutionary algorithm easily finds strategies that exploit weaknesses in the game or in opponent strategies. A method to potentially circumvent such problems is \emph{competitive coevolution} \cite{rosin1997new,stanley2004competitive,cardona2013competitive}, in which the fitness of one AI controlled player depends on its performance when competing against another player drawn from the same population of from another population. The idea here is that the evolutionary process will supply itself with worthy opponents; in the beginning of an evolutionary run, players are tested against equally bad players, but as better players develop they will play against other players of similar skill. Thus, competitive coevolution can be said to perform a kind of automated incrementalisation of the problem; ideally, this would lead to \emph{open-ended} evolution, where the sophistication and performance of evolved agents would continue increasing indefinitely due to an \emph{arms race}~\cite{rosin1997new,nolfi1998coevolving}. While open-ended evolution has never been achieved due to a number of phenomena such as cycling and loss of gradient, competitive coevolution can sometimes be very effective. 

One of the earliest examples of coevolution in games was performed by \citet{lubberts:geccoworkshop01}. The authors coevolve networks for playing a simplified version of Go and show that coevolution can speed up the evolutionary search. These results have since been corroborated by other authors~\cite{runarsson2005coevolution}. In a powerful demonstration of coevolution Blondie24 \cite{fogel2001blondie24} reached expert-level by only playing against itself. In \citet{khan2009evolution} work, the authors showed that high performing ANNs evolve quicker through coevolution than through the evaluation of agents against a Minimax-based checkers player.
%, an evolved checkers-playing ANN, 

In another example, \citet{cardona2013competitive} applied coevolutionary methods to coevolve the controllers for Ms. Pac Man and Ghost team controllers in Ms. Pac-Man. Interestingly, the authors discovered that it was significantly easier to coevolve controllers for Pac-man than for the team of ghosts, indicating that the success of a coevolutionary approach is very much dependent on the chosen domain and fitness transitivity (i.e.\ how much a solution's performance over one opponent correlates with its performance over other opponents). 

In competitive coevolution, the opponents can be drawn from the same population, or from one or several other populations. This was investigated in a series of experiments in evolving car racing controllers, where it was found that using multiple populations improved results notably~\cite{togelius2007multi}.

Coevolution does not have to be all about vanquishing one's opponents -- there is also cooperative coevolution. Whereas in competitive coevolution fitness is negatively affected by the success of other individuals, in cooperative coevolution it is positively effected. Generally, an individual's fitness is defined by its performance in collaboration with other individuals. This can be implemented on a neuronal level, where every individual is a single neuron or synapse and its fitness is the average performance of the several networks it participates in. The CoSyNE neuroevolution algorithm, which is based on this idea, was shown to outperform all other methods for variants of the pole balancing problem, a classic reinforcement learning benchmark~\cite{gomez2008accelerated}. In another example, \citet{cardamone2010applying} demonstrated that CoSyNE is also able to create competitive racing controllers for the TORCS Endurance World Championship.

The conceptually closely related algorithms ESP and SANE have been used to evolve to strategies for Othello~\cite{moriarty1995discovering} and for the strategy game \emph{Legion II}~\cite{bryant2003neuroevolution}. In another example, \citet{MLJ05} showed that coevolution can successfully discover complex control tasks for soccer Keepaway players when provided with an adequate task decomposition. However, if the problem is made more complex and the correct task composition is not given, coevolution fails to discover high performing solutions.

For some problems, more than one fitness dimension need to be taken into account. This could be the case when there is no single performance criterion (for example an open-world game with no scoring or several different scores) or when there is a good performance indicator, but evolution is helped by taking other factors into account as well (for example, you want to encourage exploration of the environment). Further, ``performance'' might just be one of the criteria you are evolving your agent towards. You might be equally interested in human-likeness, believability or variety in agent behavior. 
There are several approaches that can be taken towards the existence of multiple fitness criteria. The simplest approach is just summing the different fitness functions into one; this has the drawback that it in practice leads to very unequal selection pressure in each of the fitness dimensions. Another approach is \emph{cascading elitism}, where each generation contains separate selection events for each fitness function, ensuring equal selection pressure~\cite{togelius2007towards}.

A more principled solution to the existence of (or need for) multiple fitness functions is to use a \emph{multiobjective evolutionary algorithm} (MOEA)~\cite{fonseca1993genetic,coello2002evolutionary,deb2002fast}. Such algorithms try to satisfy all their given objectives (fitness functions), and in the cases where they partially conflict, map out the conflicts between the objectives. This map can then be used to find the solutions that make the most appealing tradeoffs between agents. These solutions, in which none of the objectives can be improved without reducing the performance of one of the other objectives, are said to be on the \emph{Pareto Front}. Multiobjective evolution has been used to balance between pure performance and other objectives in several ways. For example,~\citet{van2009robust} used an MOEA to balance between driving well and driving in a human-like manner (similar to recorded playtraces of human drivers) in a racing game.
~\citet{agapitos2008generating}, working with a different racing game, used multiobjective evolution and several different characteristics of driving style to evolve sets of NPC drivers that had visibly different driving styles while all retaining high driving skill. \citet{schrum2008constructing} used the popular multi-objective evolutionary algorithm
NSGA-II (Non-dominated Sorting Genetic Algorithm; \cite{deb2002fast}) to simply increase game-playing performance by rewarding the various components of good game-playing behavior separately.

Another promising NE approach, which might actually enable new kinds of games, is to allow a human player to interact with evolution by explicitly setting objectives during the evolutionary process, or even to act as a fitness function him or herself. Such approaches are known as {\it interactive evolution}. In the NERO video game \cite{stanley2005real}, the player can train a team of NPCs for military combat by designing an evolutionary curriculum. This curriculum consists of a sequence of training exercises that can become increasingly difficult. For example, to train the team to perform general maze navigation the player can design increasingly complex wall configurations. Once the training of the agents is complete they can be loaded into battle and tested against teams trained by other players. 

More recently, \citet{karpov:gecco11} investigated three different ways of assisting neuroevolution through human input in OpenNERO (an open-source platform based on the original NERO). One approach was the \emph{advice} method, in which users write short examples of code that are automatically converted into a partial ANN; these networks are then spliced into the evolving population. The other approach was \emph{shaping},  in which users can influence the training by modifying the environment, similar to the setup in NERO~\cite{stanley2005real}. The last approach was \emph{demonstration}, in which users can control the agent manually and the recorded data is used to train the evolving networks in a supervised fashion. The authors showed that the three ways of human-assisted NE outperform unassisted NE in different tasks, further demonstrating the promise of combining NE with human expertise.

Another example of interactive evolution in games is Galactic Arms Race (GAR~\cite{hastings2009automatic}), in which the players can discover unique particle system weapons that are evolved by NEAT. The 
fitness of a weapon in GAR is determined based on the numbers of times it was fired, allowing the players to implicitly drive evolution towards content they prefer. In Petalz \cite{risi2012combining,risi12015petalz}, a casual social game on Facebook, the core game mechanic is breeding different flowers. The player can interact with evolution by deciding which flowers to pollinate (mutation) or cross-pollinate (crossover) with other flowers.

%http://nn.cs.utexas.edu/keyword?karpov:gecco11) 

Other interactive evolution approaches to game-like environments include the evolution of two-dimensional pictures on Picbreeder~\cite{secretan2011picbreeder}, three-dimensional forms~\cite{clune11alife}, musical compositions~\cite{hoover:iccc12}, or even dance moves~\cite{dubbin2010learning}.

\section{Input Representation}
\label{sec:input_rep}

So far, we have discussed the role of neuroevolution in a game, types of neural networks, NE algorithms and fitness evaluation. Another interesting dimension by which to differentiate different uses of neuroevolution in games is what sort of input the network gets and how it is represented.  Choosing the ``right'' representation can significantly influence the ability of an algorithm to learn autonomously. This is particularly important when the role of the neural network is to control an NPC in some way. For example, in a FPS game the world could be represented to the ANN as raw sensory data, as $x$ and $y$ coordinates of various in-game characters, as distances and angles to other players and objects, or as simulated sensor data. In Ms. Pac-Man the inputs could for example be the shortest distances to all ghosts or more high-level features such as the most likely path of the ghosts~\cite{burrow2009evolution}. 

Additionally, which representation is appropriate depends to some extent on the type and on the role of the neural network in controlling the agent (see Section~\ref{sec:role}). Some representations might be more appropriate for games with direct action selection, while other representations are preferable for a board or strategy game with state evaluation. In general, the type of input representation can significantly bias the evolutionary search and dictate which strategies and behaviors can ultimately be discovered.

In this section we cover a number of ways information about the game state can be conveyed to a neural network: as straight line sensors/pie slice sensors, pathfinding sensors, third-person data, and raw sensory data. Regardless of which representation is used, there are a number of important considerations. One of the most important is that all inputs need to be scaled to be within the same range, preferably within the range $[-1, 1]$. Another is that adding irrelevant inputs can be actively harmful to performance; while in principle neuroevolution should be capable of sorting useful from useless inputs on its own, drastic performance increases can sometimes be noted when removing information from the input vector~\cite{togelius2008countering}.  

\subsection{Straight Line Sensors and Pie Slice Sensors} 
A good domain to exemplify the effect of different input representation is the simple car racing game studied by \citet{togelius2005evolving,togelius2006evolving} and \citet{togelius2007multi}. \citet{togelius2005evolving} investigated various sensor input representations and showed that the best results were achieved by using egocentric first-person information from rangefinder sensors, which return the distance to the nearest object in a given direction, instead of third-person information such as the car's position in the track's frame of reference. The authors suggest that a mapping from third-person spatial information to appropriate first-person actions is likely very non-linear, thus making it harder for evolution to discover such a controller. (These results might also be explained by the underlying fractured decision space of such a mapping~\cite{kohl2009evolving}; see Section~\ref{sec:evolving_anns}). 

In addition to evolving appropriate control policies, NE approaches can also support automatic \emph{feature selection}. \citet{whiteson2005automatic} evolved ANNs in a car racing domain and showed that an extension to NEAT allows the algorithm to automatically determine an appropriate set of straight line sensors, eliminating redundant inputs.

Rangefinder sensors are also popular in other domains like FPS games. In NERO \cite{stanley2005real} each agent has rangefinder sensors to determine their distance from other objects and walls. In addition to rangefinder sensors, agents in NERO also have ``pie slice sensors'' for enemies. These are radar-like sensors which divide the 360 degrees around the agent in a predetermined number of slices. The ANN has inputs for each of the slices and the activation of the input is proportional to the distance of an enemy unit is within this slice. If multiple units are contained in one slice, their activations are summed. Similar sensors can be found in e.g.~\citet{van2009hierarchical}. While pie slice sensors are useful to detect discrete objects like enemies or other team members, rangefinder sensors are useful to detect long contiguous objects like walls.

\subsection{Angle sensors and relative position sensors}

Another kind of egocentric sensor is the angle sensor. This simply reports the angle to a particular object, or the nearest of some class of object. In the previously discussed car racing experiments, such sensors were used for sensing waypoints which were regularly spaced out on the track~\cite{togelius2005evolving,togelius2006evolving,togelius2007multi}. In their experiment with evolving combat bots for Quake III,~\citet{Zanetti2004} used angle sensors for positions of enemies and weapon pick-ups.

Related to this are relative position sensors, that report distances to some object along some pre-specified axes. For example, \citet{yannakakis2004evolving} evolved ANNs to control the behaviour of cooperating ghosts for a clone of Pac-Man. In their setup each ANN receives the relative position to Pac-Man as input and the relative position of the closest ghost (specified by the distance along the $x$ and $y$-axis). 

\subsection{Pathfinding Sensors}
A type of sensor that is still considered egocentric but which does not take the orientation of the controlled agent into consideration, and instead takes the topology of the environment into consideration, is the pathfinding sensor. A pathfinding sensor reports the distance to the closest of some type of entity along the shortest path, as found with e.g.\ $A*$. (This distance is typically, but not always, longer than the distance along a straight line.) This is commonly used in 2D games. To take another Pac-Man example, Lucas evolved neural networks to play Ms. Pac-Man using distances along the shortest path to each ghost and to the nearest pill and power pill~\cite{lucas2005evolving}. In his case, the controller was used as a state evaluator and the actual action selection was done using one-ply search.

Different input representations can also highly bias the evolutionary search and the type of strategies that can be discovered. For example, most approaches to learning Ms. Pac Man make a distinction between edible and threat ghosts. There is typically one sensor indicating the distance to the next edible ghost, being accompanied by a similar sensor for the closest threat ghost. The idea behind this separation is to make it easier for the ANN to evolve separate  strategies to dealing with threatening and edible ghosts. While such a division makes it easier for traditional approaches to learn the game, they require domain specific knowledge that might not be available in all domains (e.g.\ edible ghosts are good, threat ghosts are bad) and might prevent certain strategies from emerging. For example, \citet{schrum2:gecco14} who used the same sensors (i.e.\ unbiased sensors) for edible and threat ghosts, showed that a modular architecture (see section~\ref{sec:ann_types}) allowed evolution to find unexpected behavioral divisions on its own. For example, a particular interesting behavior was that of luring ghosts near power pills, which would be harder to evolve with biased sensors. This example shows that there is a important relation between the evolutionary method and the type of sensors that it supports.

\subsection{Third-person Input}

In many games the evolved controller receives additional input beyond its first person sensors that is not tied to a specific frame of reference. For example, in games like Ms. Pac Man the controller typically receives the number of remaining normal and power pills or the number of edible/threat ghost \cite{schrum2:gecco14}. Another example is including the current level of damage of the car in car racing game controllers~\cite{loiacono2008wcci,loiacono20102009,cardamone2009line}.

In board games, in which the ANN does not directly control an agent (Section~\ref{sec:direct_action}), the neural network typically only receives third-person input. This could include such quantities as the piece difference and the type of piece occupying each square on the board in games like Chess~\cite{fogel2001blondie24}, Checkers~\cite{chellapilla2001evolving} or Go \cite{lubberts:geccoworkshop01}. An important aspect in this context is the \emph{geometry} of the particular domain. For example, by understanding the concept of adjacency in a game like Checkers (e.g.\ an opponent's piece can be captured by jumping over it into an unoccupied  field), a controller can learn a general strategy rather than an action tied to a specific  position on the board. While earlier attempts at evolving state evaluators did not directly take geometry into account, by representing each board position with two separate  neurons in the game Go \cite{lubberts:geccoworkshop01}, Fogel et al. \cite{fogel2001blondie24} showed that learning geometry is critical to generalization. This is true for Checkers as well. In Blondie24 each node in the first hidden layer of the ANN receives input from a different subsquare of the Checkers board (e.g.\ the first hidden node receives input from the top/right 3$\times$3 subsquare of the board). The evolutionary algorithm evolves the connection weights from the board inputs to the actual subsquares and also between the inputs and the final output node. The idea behind representing the board state as a series of overlapping subsquares is to make it easier for the ANN to discover independent local relationships within each subsquare that can then be combined in the higher levels of the network. The geometry can also be represented by a convolutional recurrent network that ``scans'' the board~\cite{schaul2009scalable}.
%their Checkers-playing neural network called

\subsection{Learning from raw sensory data}
\label{sec:rawsensory}
An exciting promise for NE approaches is to learn directly from raw sensory data instead of low-dimensional and pre-processed information. This is interesting for several reasons. One is that it might help us understand what aspects of the game's visual space is actually important, and how it should be processed, through a form of ludic computational cognitive science. Another is that forcing games to only rely on the very same information the human player gets makes for a more fair comparison with the human player, and might lead to human-like agent behaviour. More speculatively, forcing controllers to use representations that are independent from the game itself might enable more general game playing skills to develop. 

Early steps towards learning from less processed data were performed by \citet{gallagher2007evolving} in a simplified version of Pac-Man. In their approach the world was represented as a square centered around Pac-Man and the direct encoded weights of the network were optimized by evolutionary strategy. While their result demonstrated that it is possible to learn from raw data, their evolved controllers performed worse than in the experiment by \citet{lucas2005evolving} in which the shortest path distances from Pac-Man's current location to each ghost, the nearest maze junction and nearest power pill were given to the controller.

A similar setup in the game Super Mario was chosen by \citet{togelius2009super}. In their setup the authors used two grid-like sensors to detect environmental obstacles and another one for enemies. If an obstacle or an enemy occupies one of the sensors, the corresponding input to the ANN would be set to 1.0. \citet{togelius2009super} compared setups with 9 (3$\times$3), 25 (5$\times$5), 49 (7$\times$7) sensors. The authors compared a HyperNEAT-like approach with a MLP-based controller and showed that the MLP-based controller performs best with the smallest sensory setup (21 inputs total). While larger setups should potentially provide more information about the environment, a direct encoding can apparently not deal with the increased dimensionality of the search space. The HyperNEAT-like approach, on the other hand, performs equally well regardless of the size of the input window and can scale to 101 inputs because it can take the regularities in the environment description into account. The results of \citet{lucas2005evolving} and \citet{togelius2009super} suggest that there is an intricate relationship between the method and the number of sensors and the type of game they can be applied to.

HyperNEAT has also shown promise in learning from less processed data in a simplified version of the RoboCup soccer game called Keepaway \cite{verbancsics2010evolving}. Using a two-dimensional \emph{bird's eye view} representation of the game, the authors showed that their approach was able to hold the ball longer than any previously reported results from  TD-based methods like SARSA or NE methods like NEAT. Additionally the authors showed that the introduced bird's eye view representation allows changing the number of players without changing the underlying representation, enabling task transfer from 3~vs.~2 to 4~vs.~3 Keepaway without further training.

More recently \citet{TCIAIG13-mhauskn} compared how different NE methods (e.g.\ CMA-ES, NEAT, HyperNEAT) can deal with different input representation for general Atari 2600 game playing. The highest level and least general representation was called object representation, in which an algorithm would automatically identify game objects at runtime (based on object images manually identified a priori). The location of these entities was then directly provided to the evolving neural network. Similar to the work by \citet{togelius2009super}, the results by \citet{TCIAIG13-mhauskn} also indicate that while direct network encodings work best on compact and pre-processed object state representations, indirect-encodings like HyperNEAT are capable of learning directly from high-dimensional raw input data. 

While learning from raw sensory data is challenging in two-dimensional games, it becomes even more challenging in three-dimensions. One of the reasons is the need for some kind of depth perception or other distance estimation, the other is the non-locality of perception: the whole changes when you look around. In one of the early experiments to learn from raw sensory data in a three-dimensional setting, \citet{floreano2004coevolution} used a direct encoding to evolve neural networks for a simulated car racing task. In their setup, the neural network receives first-person visual input from a driving simulator called Carworld. The network is able to perform active vision through two output neurons that allow it to determine its visual focus and resolution in the next time step. While the actual input to the ANN was limited to 5$\times$5 pixels of the visual field, the evolved network was able to drive better or equal to well-trained human drivers. Active vision approaches have also been successfully  applied to board games like Go, in which a ``roving eye'' can self-directedly traverse the board and stop where it thinks the next stone should be placed~\cite{stanley:gecco04}. More recently, \citet{koutnik2013evolving} evolved an indirectly encoded and recurrent controller for car driving in TORCS, which learned to drive based on a raw 64$\times$64 pixel image. That experiment featured an indirect encoding of weights analogous to the JPEG compression in images.

\citet{parker2008neuro, parker2012neurovisual} evolved an ANN to shoot a moving enemy in Quake II, by only using raw sensory information. In their setup the bot was restricted to a flat world in which only a band of 14$\times$2 gray-scale pixels from the center of the screen was used as input to the network. However, although the network learned to attack the enemy, the evolved bots were shooting constantly and spinning around in circles, just slowing down minimally when an enemy appeared in their view field. While promising, the results are still far from the level of a human player or indeed one that has received more processed information such as angles and distances.

\section{Open Challenges}
\label{sec:outlook}

This paper has presented and categorized a large body of work where NE has been applied, mostly but not exclusively in the role of controlling a game agent or representing a strategy. There are many successes, and NE is already a technique that can be applied more or less out of the box for some problems. But there are also some domains and problems where we have not yet reached satisfactory performance, and other tasks that have not been attempted. There are also various NE approaches that have been only superficially explored. In the following, we list what we consider the currently most promising future research directions in NE. While there are plenty of basic research questions in evolutionary computation and NE, which are important for the application of these techniques in games, this section will mostly focus on applied research in the sense of research motivated by use in games.

\subsection{Reaching Record-beating Performance}
We have seen throughout the paper that NE performs very well in many domains, especially those involving some kind of continuous control. For some problems (see Section~\ref{sec:recordbeating}) NE is the currently best known method. Extending the range of problems and domains on which NE performs well is an important research direction in its own right. In recent years, Monte Carlo Tree Search (MCTS) has provided record-beating performance in many game domains~\cite{browne2012survey}. It is likely that many clues can be taken from this new family of algorithms in order to improve NE, and there are probably also hybridisations possible. Of course, performance can be measured in many different ways; in game tasks, it is often (though not always) about playing well (for measure of good playing) within some given computation time limit.

\subsection{Comparing and combining evolution with other learning methods}
\label{section:comparing}

While NE is easily applicable and often high-performing, and sometimes the best approach available for some problem, it is almost never the only type of algorithm that can be applied to a problem. There are always other evolvable representations, such as expression trees used in genetic programming. For player modeling, supervised learning algorithms based on gradient descent can often be applied, and for reinforcement learning problems, one could choose to apply algorithms from the temporal difference learning family (e.g.\ TD(0) or Q-learning). The relative performance of alternative methods compared to Q-learning differs drastically; sometimes NE really is the best thing to use, sometimes not. The outstanding research question here is: when should one use NE?

From the few published comparative studies of NE with other kinds of reinforcement learning algorithms, we can learn that in many cases, TD-based algorithms learn faster but are more brittle and NE eventually reaches higher performance~\cite{lucas2007point,runarsson2005coevolution,whiteson2007empirical}. But sometimes, TD-learning performs very well when well tuned~\cite{lucas2006temporal}, and sometimes NE completely dominates all other algorithms~\cite{gomez2008accelerated}. What is needed is some sort of general theory of what problem characteristics advantage and disadvantage NE; for this, we need parameterizable benchmarks that will help us chart the problem space from the vantage point of algorithm performance~\cite{togelius2009ontogenetic}. There have been some attempts at constructing such benchmarks previously~\cite{kalyanakrishnan2011characterizing}, and the General Video Game Playing Competition characterizes its games according to game design characteristics (problem features), allowing another way of comparing performance on problem classes~\cite{perez201522014}.

But mapping the relative strengths of these algorithms is really just the first step. Once we know when NE works better or worse than other algorithms, we can start inventing hybrid algorithms that combine the strengths of both neuroevolution and its alternatives, in particular TD-learning and GP. Previous research by Whiteson and Stone in combining NEAT and Q-learning has shown promising results \cite{whiteson:jmlr2006,whiteson2006line}. These methods have been applied with some success to shooters~\cite{reeder2008interactively} and racing games~\cite{cardamone2010learning}.

\subsection{Learning from high-dimensional/raw data}

As discussed in Section~\ref{sec:rawsensory}, learning from raw images or similar high-dimensional unprocessed data is a hard challenge of considerable scientific interest and with several applications. The paucity of experiments in evolving neural networks that control game agents based on raw data is puzzling given the fertility of this topic. However, as we can see from the published results, it is also very hard to make this work. It stands to reason why the best results have been achieved using drastically scaled down first-person image feeds. Direct shallow approaches seem to be unable to deal with the dimensionality and signal transformation necessary for extracting high-level information from high-dimensional feeds. However, recent advances in convolutional networks and other deep learning architectures on one hand, and in indirect encodings like HyperNEAT on the other, promise significantly improved performance. It seems like very interesting work would result from the sustained application of these techniques to e.g.\ the visual inputs from Quake II. This kind of task might also be the catalyst for the development of new evolvable indirect neural network encodings.

\subsection{General video game playing}
\label{gvgp}

One of the strengths of NE is how generic it is; the same algorithm can, with relatively few tweaks, be applied to a large number of different game-related tasks. Yet, almost all of the papers cited in this paper use only a single game each for a testbed. The problem of how to construct NE solutions that can learn to play any of a number of games is seriously understudied. One of the very few exceptions is the recent work on learning to play arbitrary Atari games~\cite{TCIAIG13-mhauskn,hausknecht:gecco12}. While NE performed admirably in that work, there is clearly scope for improvement. Apart from the Atari GGP benchmark, another relevant benchmark here is the General Video Game Playing Competition, which uses games encoded in the Video Game Description Language (VGDL)~\cite{ebner2013towards,schaul2013video}. This means that unlike the Atari GGP benchmark, this competition (and its associated benchmark software) can feature generated games, and thus a theoretically unbounded set of video games. A controller architecture that could be evolved to play any game from a large set would be a step towards more generic AI capabilities. The first edition of the competition was won by controllers based on variations of Monte Carlo Tree Search, but future editions will feature a ``learning track'' which allows controllers considerable time to train on each game~\cite{perez201522014} -- NE methods are likely to be competitive here.

\subsection{Combining NE with life-long learning} %Combining NE with life-long learning
\label{sec:ne_life_long}
An even larger step would to be evolve a single neural network that could \emph{learn} and \emph{adapt} during its lifetime (ontogenetically, i.e.\ without further evolution) to play one of a set of games. This has as far as we know never been attempted, but would be highly impressive. Evolution and learning are two forms of biological adaptation that operate on very different timescales. Learning can allow an organism to adapt much faster to environmental changes by modifying its behaviors during its lifetime. One way that NE can create such \emph{adaptive ANNs} is to not only evolve the weights of an ANN but also local synaptic plasticity parameters that determine how the weights of the network change during the \emph{lifetime} of the agent based on incoming activation \cite{soltoggioDuerrMattiussiFloreano2007,urzelaiFloreano2001adaptive,floreano2008neuroevolution,tonelli2013relationships}. This resembles the way the brains of organisms in nature can cope with changing and unpredictable situations \cite{hebb1949}.     

While there has been progress in this field, adaptive ANNs have so far mostly been applied to relatively simple toy problems. However, novel combinations of recent advances such as more advanced forms of local plasticity (e.g.\ neuromodulation~\cite{soltoggioDuerrMattiussiFloreano2007}), hypothesis testing in distal reward learning \cite{soltoggio2015plasticity}, larger indirectly-encoded adaptive networks \cite{risi2012unified,risi2010indirectly,risiStanley2014}, methods that avoid deception inherent in evolving learning architectures \cite{lehman:gecco14,risi2010evolving}, and learning of large behavioral repertoires \cite{cully2015robots}, could allow the creating of learning networks for more complex domains such as games.

Such adaptive networks could overcome many of the challenges in applying NE to games, such as adjusting on the fly to the difficulty of the opponent, incrementally learning new skills without forgetting current ones, and ultimately allow general video game playing ANNs. However, preventing these networks from potentially learning undesired behaviors in addition to being reliable and controllable, are important future research direction, especially in the context of commercial games (Section~\ref{sec:commercial}).

\subsection{Competitive and cooperative coevolution}

In our discussion of fitness functions in Section~\ref{sec:fitness}, we discuss competitive and cooperative coevolution at some length. This is because these approaches bear exceptional promise. Competitive coevolution could in theory enable open-ended evolution through arms races; cooperative coevolution could help find solutions to complex problems through automatically decomposing the problems and evaluating partial solutions by how well they fit together. Unfortunately, various problems beset these approaches and prevent them from achieving their full potential. For cooperative coevolution there is the problem of how to select the appropriate level of modularisation, i.e.\ which are the units that are cooperatively coevolved. Competitive coevolution has several pathologies, such as cycling and loss of gradient. However, we suspect that these problems have as much to do with the benchmark as with the algorithm. For example, open-ended evolution might not be achievable in the predator-prey scenarios that were used in previous research, as there is just no room for more sophisticated strategies. Modern games might provide the kind of environments that would allow more open-ended evolution to take place.

\subsection{Fast and reliable methods for commercial games}
\label{sec:commercial}
This paper has been an overview of the academic literature on NE in games rather than of the uses of NE in the game industry, for the simple reason that we do not know of many examples of NE being part of published commercial games (with the exception of the commercial game \emph{Creatures}~\cite{grand:agents97} and indie titles such as GAR~\cite{hastings2009automatic} and Petalz~\cite{risi2012combining,risi2014automatically}). Therefore, one key research problem is to identify which aspects of neural networks, evolutionary algorithms and their combination have hindered its uptake in commercial game AI and try to remedy this. For example,  ANNs have mostly found their way into commercial games for data mining purposes, with a few exceptions of usage for NPC control in games such as \emph{Black$\&$White} or the  car racing game \emph{Colin McRae Rally 2}. Game developers often cite the lack of control and clarity as an issue when working with neural networks. Especially if the ANN can learn \emph{on-line} while the game is being played (Section~\ref{sec:ne_life_long}), how can we make sure it does not suddenly kill an NPC character that is vital to the game's story? Additionally, if the NPCs can change their behavior, game balancing is more challenging and new types of debugging tools might become necessary. In the future, it will be important to 
address these challenges to encourage the wider adoption of promising NE techniques in the game industry.

\section*{Acknowledgements}
We thank the numerous colleagues who have graciously read and commented on versions of this paper, including Kenneth O. Stanley, Julian Miller, Matt Taylor, Mark J. Nelson, Siang Yew Chong, Shimon Whiteson, Peter J. Bentley, Jeff Clune, Simon M. Lucas, Peter Stone and Olivier Delalleau.

\bibliographystyle{abbrvnat}

\bibliography{neuroevolution-in-games}

\end{document}